\documentclass[10pt,twocolumn,letterpaper]{article}
\usepackage[textwidth=7in,textheight=9in,columnsep=0.375in,centering]{geometry}
\usepackage[T1]{fontenc}
\usepackage{newtxtext,newtxmath}
\usepackage[hyphens]{url}
\usepackage{graphicx}
\usepackage[round]{natbib}
\usepackage{caption}
\usepackage{booktabs}
\usepackage{needspace}
\usepackage{placeins}
\usepackage{xspace}
\usepackage{amsmath}
\usepackage[table]{xcolor}
\usepackage{colortbl}
\usepackage[colorlinks=true,linkcolor=blue!60!black,citecolor=blue!60!black,urlcolor=blue!60!black]{hyperref}
\AtBeginDocument{\let\cite\citep}

\makeatletter
\renewcommand\normalsize{\@setfontsize\normalsize\@xpt{11}\abovedisplayskip 7pt plus 2pt minus 4pt\belowdisplayskip\abovedisplayskip\abovedisplayshortskip 0pt plus 2pt\belowdisplayshortskip 4pt plus 2pt minus 2pt}
\renewcommand\small{\@setfontsize\small\@ixpt{10}}
\renewcommand\footnotesize{\@setfontsize\footnotesize\@ixpt{10}}
\renewcommand\section{\@startsection{section}{1}{\z@}{-2.0ex plus -0.5ex minus -.2ex}{3pt plus 2pt minus 1pt}{\large\bfseries}}
\renewcommand\subsection{\@startsection{subsection}{2}{\z@}{-2.0ex plus -0.5ex minus -.2ex}{3pt plus 2pt minus 1pt}{\normalsize\bfseries}}
\makeatother
\normalsize
\definecolor{boarow}{rgb}{0.90,0.95,1.0}
\newcommand{\methodName}{Boa\xspace}

\hypersetup{pdftitle={Linear-Time Context Compression with State-Space Encoding},pdfauthor={Marc Serra Ortega, Andrey Barsky, Carlos Boned Riera, Tom\'{a}s Ockier, Dimosthenis Karatzas},pdfsubject={},pdfkeywords={},pdfcreator={},pdfproducer={}}

\title{Linear-Time Context Compression with State-Space Encoding}
\author{Marc Serra Ortega\thanks{Corresponding author: \href{mailto:mserrao@cvc.uab.cat}{mserrao@cvc.uab.cat}} \qquad Andrey Barsky \qquad Carlos Boned Riera\\
Tom\'{a}s Ockier \qquad Dimosthenis Karatzas\\[0.5em]
Computer Vision Center, UAB, Spain}
\date{}

\begin{document}
\maketitle

\begin{abstract}
Transformer-based large language models (LLMs) impose substantial computation and memory costs as context sizes increase, motivating methods that compress a context into a shorter representation before inference. Existing soft-prompt compression methods typically use the same Transformer backbone as used in the target model, where producing the compressed representation leads to significant latency overheads. This restricts them to the offline setting where context is compressed once and then reused across many queries.
We propose the \textbf{\methodName Compressor}, a plug-and-play \textit{online} context compressor built on a Mamba state-space model (SSM) that encodes arbitrary-length context in a single linear-time pass, producing a representation on which we condition a frozen target LLM. The SSM is trained by distillation, matching the next-token logit distribution from the LLM ingesting the full context (as teacher) to the same LLM ingesting the compressed representation (as the student).
At equal compression ratios, \methodName matches or surpasses the performance of existing state-of-the-art offline compression baselines, while being an order of magnitude faster than Transformer-based implementations, thereby \textit{reducing} end-to-end latency rather than adding compression time overhead. We demonstrate that \methodName also scales well to longer contexts and to compression ratios as aggressive as 16$\times$.
By enabling effective compression that is cheap enough to run on every query at inference time, \methodName becomes a practical and general tool for efficient long-context inference.

\end{abstract}

\section{Introduction}
\label{sec:intro}

\begin{figure}[!t]
\centering
\includegraphics[width=0.95\columnwidth]{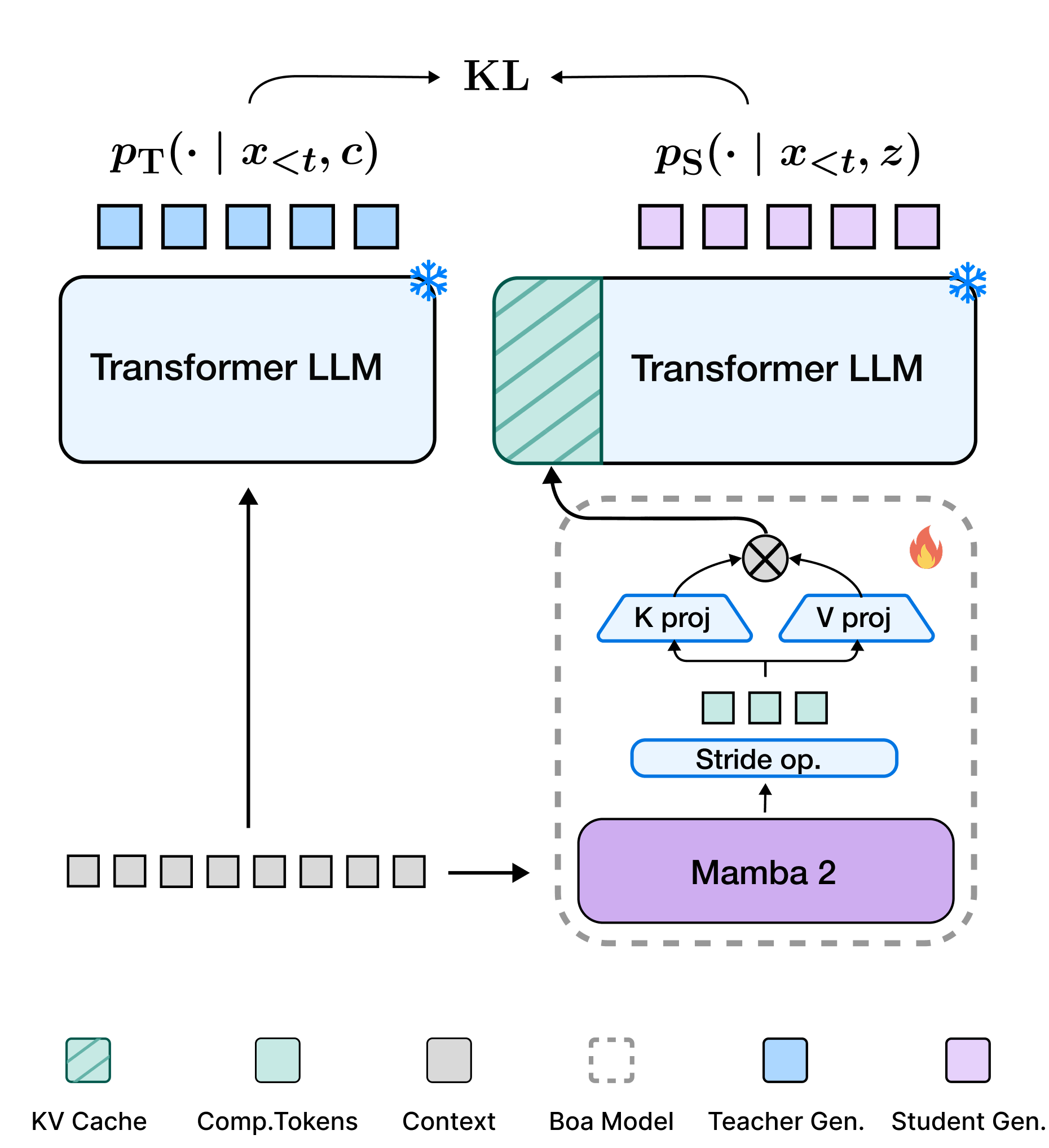}
\caption{Overview of the Boa compressor. A trainable Boa module encodes the input context using a Mamba 2 model, a temporal stride operation, and key-value (KV) projections. This compressed KV cache is injected into a frozen student Transformer \emph{decoder}. The Boa \emph{encoder} is trained by minimising the KL divergence between the student's output and that of a frozen teacher model.}
\label{fig:method}
\end{figure}

Large language models are increasingly deployed on long inputs, including multi-document question answering, retrieval-augmented generation, agent trajectories, and long-form reasoning. However, the Transformer architecture that enables these models has scaling issues at long contexts: self-attention costs grow quadratically with sequence length during pre-fill, and the key-value (KV) cache grows linearly during decoding. With longer inputs, inference cost is increasingly dominated by these time and memory constraints.

Context compression is an answer to this scaling problem: replacing a long context with a compact representation that preserves the information the model needs to answer. Effective compression simultaneously reduces pre-fill cost and cache size, filters noise from long inputs \cite{jiang2023longllmlingua}, and extends the effective context size that a model can handle.
The highest information per token can be retained using \emph{soft-prompt} compression methods, which learn to encode the context into a short sequence of continuous vectors that condition a frozen decoder. Recent works~\cite{zhang2024activation, ge2024icae, li2024500x, liu2026sac} have demonstrated that much of the performance of the model can be retained even at high compression ratios.

However, existing methods share a limitation in their choice of compressor architecture. In most cases, the compressor uses the same Transformer architecture as the target model: that is, the encoder is the LLM itself, or an adaptation of it, or otherwise another Transformer of comparable size.
In practice this restricts soft-prompt compression to the offline regime, where a context is compressed once and reused across many queries. This is useful in a range of circumstances, but for a context seen only once (a common case at inference time), the cost of compression is comparable to processing the original context itself.

This limitation is not inherent to soft-prompt compression methods, but only to the choice of compressor architecture.
State-space models (SSMs) such as Mamba \cite{gu2024mamba} process a sequence in linear time with a fixed-size recurrent state, while remaining competitive with Transformers on long-sequence modelling. This class of models makes for natural context compressors: they can read an arbitrarily long input cheaply and summarise it into a compact representation, offloading the heavy reasoning to a frozen Transformer decoder.
This makes compression cheap enough to run \emph{online}, as a single-pass step in the inference path itself, rather than only as \emph{offline} preprocessing.

We instantiate this idea in the \methodName Compressor (\figurename~\ref{fig:method}). A Mamba encoder reads the context and produces a short sequence of representations that are injected directly into a frozen LLM; the LLM is never modified or fine-tuned. \methodName is trained by knowledge distillation, matching the frozen model's own output distribution on the full context, so the compressed representation is optimised to directly reproduce the behaviour of the decoder. On the MRQA benchmark suite \cite{fisch-etal-2019-mrqa}, \methodName matches or surpasses state-of-the-art Transformer-based baselines at equal compression ratios, and degrades far more gracefully as the ratio increases, while cutting total long-context latency by more than an order of magnitude.

The remainder of the paper reviews related work in context compression and state-space models (Section~\ref{sec:rw}), details the \methodName architecture and training (Section~\ref{sec:method}), reports quality and efficiency results (Section~\ref{sec:exp}), and discusses broader implications on compressor design (Section~\ref{sec:disc}).

\section{Related Work}
\label{sec:rw}

\subsection{Context Compression}
\label{sec:rw-cc}

The objective of compressing an LLM's input context is to reduce the number of tokens it must process while generating output sufficiently similar to that attained without compression. There are broadly two families of approach to this: \emph{hard-prompt} methods that delete or rewrite tokens in natural language, and \emph{soft-prompt} methods that replace tokens with continuous representations.

\textbf{Hard-prompt compression} methods keep the input in natural language and remove what is redundant. Selective-Context \cite{selectivecontext} prunes low-information spans using token-level entropy or perplexity, and the LLMLingua family \cite{jiang2023llmlingua,jiang2023longllmlingua} extends this to aggressive token deletion, showing that models can act on inputs even when fluency is broken. These methods are model-agnostic and require no training, but they are bounded by what can be expressed in surviving tokens and cannot compress below the granularity of the text itself.

\textbf{Soft-prompt compression} methods compress natural-language tokens into a small number of continuous vectors, achieving higher density than hard prompts.  Gisting \cite{mu2023learning} trains reusable ``gist'' tokens, though it requires fine-tuning the target model and handles only short prompts.
AutoCompressor \cite{chevalier2023autocompressor} scales input length by recurrently propagating summary vectors across segments; however, it too requires fine-tuning the target model and incurs heavy encoding costs due to performing multiple forward passes over the backbone for a single context.
ICAE \cite{ge2024icae} removes the need to change the decoder: an LLM adapted with LoRA encodes a context of up to roughly 500 tokens into a handful of memory slots that condition the frozen model, reaching up to 8$\times$ compression with a two-step training procedure: first pre-trained with a combination of autoencoder and language modelling losses, then fine-tuned on instruction data.
500xCompressor \cite{li2024500x} shows that placing the compressed representation in the KV space rather than the embedding space is able to achieve far higher effective compression ratios of up to 480$\times$ under similar conditions.

Soft prompts have also been specialised for retrieval-augmented generation (RAG), where a passage is compressed once and reused, as in xRAG and COCOM \cite{cheng2024xrag,rau2024cocom}.

We compare against two state-of-the-art methods that serve as the closest comparison baselines. \textbf{Activation Beacon} \cite{zhang2024activation} compresses the KV activations produced throughout the decoder by splitting the context into chunks, summarising each chunk with special beacon tokens whose activations are retained, and training via compression-based autoregression at randomly sampled ratios.
While it preserves near-uncompressed quality up to $128\text{K}$ tokens and reduces cache memory, encoding is computationally expensive, requiring multiple forward passes over the decoder, and compression is structurally bounded by its sliding-window size.
\textbf{Semantic-Anchor Compression (SAC)} \cite{liu2026sac} removes the reconstruction-based pretraining used by ICAE-style methods, which it shows interferes with the language-modelling features needed downstream; instead it selects existing context tokens as anchors and aggregates information into them through bidirectional attention. SAC improves alignment between compressed and original representations and confirms that pretraining on a language-modelling objective, rather than autoencoding, yields stronger downstream compression.

Across this family of methods, there is an architectural assumption that the compressor is a Transformer (often adapted from the target model itself) and therefore that encoding any context carries Transformer-order computational costs.  \methodName keeps the soft-prompt idea and the frozen decoder but replaces the encoder with a linear-time SSM, so the compression step itself is computationally cheap.

\textbf{Context distillation.} Knowledge distillation (KD; \citealt{hinton2015distilling}) has been widely applied for model compression, but has also been applied to conditioning LLMs without full context: by directly optimising per-prompt soft tokens \cite{wingate2022} or internalising the context into the LLM's own weights \cite{snell2022learning}. Closer to our work is xRAG \cite{cheng2024xrag}, which trains a lightweight projection over frozen retrieval embeddings by distilling from the full-context model.
For \methodName we adopt the distillation framework to encode arbitrary contexts with a trainable sequence encoder, as a natural objective for preserving fidelity between compressed and uncompressed contexts, finding it to work better than direct language modelling.

\textbf{Online vs. offline compression. }  When a context is to be reused across several queries, it is efficient to compress it once in advance, and the cost of compression is repaid over repeated access. We call this the \textit{offline} setting. By contrast, in the \textit{online} setting, the context is seen for the first and last time at inference, so compression must pay for itself within a single query.
Transformer-based methods (most of those described above) are typically constrained to offline use, as the overhead cost of single-use compression adds latency rather than removing it.

Existing work on online compression has been largely confined to the RAG setting: RECOMP \cite{xu2024recomp} summarises passages between retrieval and encoding, and REFRAG \cite{lin2025refrag} selectively expands query-relevant retrievals based on pre-indexed encodings. LongLLMLingua \cite{jiang2023llmlingua} is a hard-prompt method that prunes tokens conditioned on the query, but is limited in performance to the granularity of the text being compressed. These methods are effective specifically when context is highly redundant and independently chunked, with sparse relevance to the query.

The online setting also offers an advantage that is unavailable offline: since the query is known at compression time, the compressor benefits from conditioning on it and retaining only the information in the context required to answer.
In practice, this advantage has not been fully exploited: existing online compression methods achieve high performance only under narrow assumptions, and offline compression methods are usually too slow to see single-query benefits (though we observe in Figure~\ref{fig:beacon} that SAC achieves moderate latency gains at longer context lengths). Boa is designed for precisely this setting, achieving the high performance of Transformer-based offline compressors, with a latency that makes it effective in the online regime.

\subsection{State-Space Models for Compression}
\label{sec:rw-ssm}

State-space models (SSMs) process sequences through a latent recurrent state, enabling linear-time inference and constant-size memory. Selective variants such as Mamba \cite{gu2024mamba} and Mamba-2 \cite{dao2024mamba2} improve earlier structured SSMs by making state transitions input-dependent.

A known limitation of SSMs relative to attention-based models is weaker associative recall and verbatim copying, due to the fixed size of the recurrent state \cite{jelassi2024repeat, arora2024zoology, waleffe2024empirical}. This motivates our use of the SSM in a hybrid architecture: \methodName does not require retrieval from the compressed state itself, but allows a Transformer full attention over the set of compressed state representations, which scales as a fraction of input length.

RCC \cite{huang2024rcc} applies SSMs to context compression, but processes the input in independent chunks, and fine-tunes both the encoder and decoder. In contrast, \methodName compresses in a single pass and remains plug-and-play by retaining a frozen LLM as the target decoder.

\begin{table*}[!t]
\centering
\footnotesize
\setlength{\tabcolsep}{3.2pt}
\renewcommand{\arraystretch}{1.2}
\begin{tabular}{l cc cc cc cc cc cc| cc | cc}
\toprule
& \multicolumn{2}{c}{SQuAD} & \multicolumn{2}{c}{NewsQA} & \multicolumn{2}{c}{TriviaQA} & \multicolumn{2}{c}{SearchQA} & \multicolumn{2}{c}{HotpotQA} & \multicolumn{2}{c}{NaturalQ} & \multicolumn{2}{c|}{Average} & \multicolumn{2}{c}{Norm~$\Delta$} \\
\cmidrule(lr){2-3}\cmidrule(lr){4-5}\cmidrule(lr){6-7}\cmidrule(lr){8-9}\cmidrule(lr){10-11}\cmidrule(lr){12-13}\cmidrule(lr){14-15}\cmidrule(lr){16-17}
Method & EM & F1 & EM & F1 & EM & F1 & EM & F1 & EM & F1 & EM & F1 & EM & F1 & EM & F1 \\
\midrule
\rowcolor{black!5}\textit{Full context (ceiling)} & 87.29 & 93.70 & 57.17 & 72.06 & 76.12 & 82.92 & 81.53 & 87.50 & 68.95 & 83.76 & 70.60 & 82.52 & 73.61 & 83.74 & \textcolor{green!60!black}{100} & \textcolor{green!60!black}{100} \\
\rowcolor{black!5}\textit{No context (floor)} & 5.70 & 12.19 & 2.04 & 4.63 & 21.70 & 24.91 & 15.18 & 17.91 & 4.07 & 9.28 & 5.97 & 11.84 & 9.11 & 13.46 & \textcolor{red!70!black}{0} & \textcolor{red!70!black}{0} \\
\midrule
\multicolumn{17}{c}{\textit{4$\times$~compression}} \\
\midrule
SAC & 64.19 &	79.63 &	44.92 &	62.39 &	64.92 &	72.69 &	\underline{80.57} &	\underline{85.88} &	61.51 &	77.9 &	\underline{61.52} &	\underline{75.25} &	\underline{62.94} &	\underline{75.62} &	\underline{83.46} &	\underline{88.45} \\
Beacon & \underline{67.07} & \underline{82.57} & \underline{48.05} & \underline{67.20} & \underline{67.53} & \underline{74.82} & 63.81 & 72.60 & \underline{63.41} & \underline{79.49} & 60.28 & 75.02 & 61.69 & 75.28 & 81.52 & 87.96 \\
\rowcolor{boarow}\textbf{\methodName (ours)} & \textbf{79.81} & \textbf{89.43} & \textbf{52.75} & \textbf{68.83} & \textbf{76.87} & \textbf{82.66} & \textbf{81.33} & \textbf{87.17} & \textbf{63.85} & \textbf{80.30} & \textbf{66.20} & \textbf{79.32} & \textbf{70.14} & \textbf{81.29} & \textbf{94.62} & \textbf{96.51} \\
\midrule
\multicolumn{17}{c}{\textit{8$\times$~compression}} \\
\midrule
SAC & 53.17 & 70.28 & 37.73 & 55.73 & 61.98 & 70.09 & \underline{79.43} & \underline{85.05} & 57.62 & 75.15 & 54.21 & 69.90 & 57.36 & 71.03 & 74.81 & 81.92 \\
Beacon & \underline{64.66} & \underline{80.66} & \underline{44.23} & \underline{64.22} & \underline{65.39} & \underline{73.01} & 61.15 & 70.03 & \underline{60.40} & \underline{77.57} & \underline{57.08} & \underline{72.59} & \underline{58.82} & \underline{73.01} & \underline{77.07} & \underline{84.73} \\
\rowcolor{boarow}\textbf{\methodName (ours)} & \textbf{78.01} & \textbf{88.40} & \textbf{52.30} & \textbf{68.38} & \textbf{76.48} & \textbf{82.34} & \textbf{81.05} & \textbf{86.84} & \textbf{62.53} & \textbf{79.20} & \textbf{63.74} & \textbf{77.58} & \textbf{69.02} & \textbf{80.46} & \textbf{92.88} & \textbf{95.33} \\
\midrule
\multicolumn{17}{c}{\textit{16$\times$~compression}} \\
\midrule
SAC & 34.01 &	51.33 &	26.45 &	43.6 &	56.08 &	64.89 &	74.28 &	80.9 &	44.48 &	62.96 &	40.6 &	57.18 &	45.98 &	60.14 &	46.27 &	66.42 \\
Beacon & \underline{60.59} & \underline{77.05} & \underline{41.57} & \underline{61.27} & \underline{62.63} & \underline{70.69} & \underline{56.83} & \underline{65.58} & \underline{56.87} & \underline{74.03} & \underline{54.37} & \underline{70.18} & \underline{55.48} & \underline{69.80} & \underline{71.89} & \underline{80.17} \\
\rowcolor{boarow}\textbf{\methodName (ours)} & \textbf{78.16} & \textbf{88.16} & \textbf{51.64} & \textbf{67.64} & \textbf{76.39} & \textbf{82.21} & \textbf{81.18} & \textbf{86.95} & \textbf{61.84} & \textbf{78.56} & \textbf{63.55} & \textbf{77.45} & \textbf{68.79} & \textbf{80.16} & \textbf{92.53} & \textbf{94.91} \\
\bottomrule
\end{tabular}
\caption{MRQA \emph{in-domain} results with Qwen3-1.7B, per dataset. EM / token-level F1; higher is better. \textbf{Norm~$\Delta$} normalises the average between the no-context floor (0) and full-context ceiling (100). \methodName{} uses the Mamba-2 (1.3B) KV variant.}
\label{tab:indomain}
\end{table*}

\section{Method}
\label{sec:method}

\subsection{Overview}
\label{sec:method-overview}

\methodName compresses a long context into a short sequence of continuous representations that condition the target frozen LLM. The method consists of a trainable Mamba-based encoder and a lightweight projection module, while the target Transformer decoder remains frozen. Given an input sequence of length $L$, the encoder processes it in a single linear-time pass. A strided subset of the resulting hidden states is then projected into the attention space of the decoder to produce the layer-wise key ($K$) and value ($V$) representations. The decoder attends to this compressed KV cache as contextual information and generates the output. \figurename~\ref{fig:method} illustrates the overall architecture.
For a compression ratio $r \in \mathbb{N}$, we use a stride $r$ and produce $m = \lceil L/r\rceil$ compressed representations.

\subsection{Mamba Compressor}
\label{sec:method-encoder}

The encoder is a pretrained Mamba2-based model that processes the input in linear time. Let the context be $c=(x_1,\dots,x_L)$ with token embeddings $(u_1,\dots,u_L)$. A selective state-space layer maintains a hidden state $h_t$ that evolves recurrently with input-dependent parameters:
\begin{equation}
\label{eq:ssm}
h_t = \bar{A}_t\,h_{t-1} + \bar{B}_t\,u_t, \qquad e_t = C_t\,h_t,
\end{equation}
where $(\bar{A}_t,\bar{B}_t,C_t)$ are functions of the input, via Mamba's selection mechanism, and $e_t$ is the encoder hidden state at position $t$. Since the state is carried forward recurrently, each $e_t$ summarises all tokens up to position $t$ in a single pass.

Rather than emit one representation per token, \methodName gathers hidden states on a fixed stride. Anchored at the final token and stepping backward by $r$, it selects the index set $\mathcal{I}=\{\,L,\,L-r,\,L-2r,\,\dots\,\}$ with $|\mathcal{I}|=m=\lceil L/r\rceil$, yielding the compressed sequence $E=(e_i)_{i\in\mathcal{I}}$. The last position is always the final token of the context, which is typically the most information-dense position in a causal model. Earlier positions are periodic checkpoints back through the context.

For our experiments, we use Mamba-2 pretrained on roughly $300$B tokens. This gives us a suitable pretrained base encoder to evaluate against prior methods.
\subsection{Conditioning the Frozen Decoder}
\label{sec:method-cond}

The compressed representations can condition the Transformer decoder in two ways. The most straightforward is to feed them as input embeddings, i.e. as soft prompts that enter at the first layer of the decoder. The stronger method, which we adopt, is to inject them directly as per-layer key and value vectors in the attention cache of the decoder. For each selected state $e_i$ and decoder layer $\ell\in\{1,\dots,N\}$, a learned projection $P^{\ell}$ produces a key/value pair:
\begin{equation}
\label{eq:proj}
(k_i^{\ell},\,v_i^{\ell}) = P^{\ell}(e_i),
\end{equation}
and the collection $Z=\{(k_i^{\ell},v_i^{\ell})\}_{i\in\mathcal{I}}^{\ell}$ is assembled into a KV cache packed exactly as one the decoder would have produced for a real prefix of length $m$, including the rotary position encoding of the model, so that from the perspective of the decoder the compressed representations occupy the first $m$ positions of its context.

KV injection conditions the model at every layer rather than only in the input embedding, giving each compressed representation substantially more capacity to influence the network than a single soft token. This mirrors the finding in 500xCompressor~\cite{li2024500x} that KV-space compression outperforms embedding-space compression at high ratios. The key difference in our setting is that the encoder of 500xCompressor, and posterior methods, is the LLM itself, so its keys and values already live in the attention space of the decoder and can be extracted by slicing the model's own cache. Our encoder is a separate Mamba, with its own distinct representation space. Therefore, the decoder keys and values cannot be obtained by directly slicing encoder activations and must instead be learned through a projection module, which serves as the interface between the two architectures.
This design is non-trivial because injecting learned key-value representations into every decoder layer modifies the attention computation throughout the network depth and may destabilise generation.
Motivated by prefix tuning and 500xCompressor, we nevertheless condition the frozen decoder directly in KV space. Our results show that the decoder can effectively exploit keys and values produced by a heterogeneous encoder, achieving performance comparable to compression methods that reuse the target model's architecture.

\begin{table*}[!t]
\centering
\footnotesize
\setlength{\tabcolsep}{3.2pt}
\renewcommand{\arraystretch}{1.2}
\begin{tabular}{l cc cc cc cc cc cc| cc | cc}
\toprule
& \multicolumn{2}{c}{BioASQ} & \multicolumn{2}{c}{DROP} & \multicolumn{2}{c}{DuoRC} & \multicolumn{2}{c}{RACE} & \multicolumn{2}{c}{RE} & \multicolumn{2}{c}{TextbookQA} & \multicolumn{2}{c|}{Average} & \multicolumn{2}{c}{Norm~$\Delta$} \\
\cmidrule(lr){2-3}\cmidrule(lr){4-5}\cmidrule(lr){6-7}\cmidrule(lr){8-9}\cmidrule(lr){10-11}\cmidrule(lr){12-13}\cmidrule(lr){14-15}\cmidrule(lr){16-17}
Method & EM & F1 & EM & F1 & EM & F1 & EM & F1 & EM & F1 & EM & F1 & EM & F1 & EM & F1 \\
\midrule
\rowcolor{black!5}\textit{Full context} & 61.30 & 77.32 & 55.49 & 64.09 & 56.63 & 66.52 & 45.55 & 59.34 & 77.82 & 88.13 & 64.80 & 74.16 & 60.27 & 71.59 & \textcolor{green!60!black}{100} & \textcolor{green!60!black}{100} \\
\rowcolor{black!5}\textit{No context} & 17.35 & 24.06 & 23.09 & 26.38 & 1.67 & 4.58 & 2.67 & 6.03 & 9.74 & 15.26 & 24.28 & 31.48 & 13.13 & 17.97 & \textcolor{red!70!black}{0} & \textcolor{red!70!black}{0} \\
\midrule
\multicolumn{17}{c}{\textit{4$\times$~compression}} \\
\midrule
SAC & 55.12 & 68.95 &	45.44 &	\textbf{54.39} &	48.57 &	58.9 &	\textbf{37.98} &	\textbf{53.31} &	\underline{68.42} &	\textbf{81.14} &	\textbf{62.28} &	\textbf{70.67} &	\textbf{52.97} &	\textbf{64.56} &	\textbf{84.51} &	\textbf{86.89} \\
Beacon & \underline{51.99} & \underline{67.99} & \underline{41.18} & \underline{51.14} & \underline{49.70} & \underline{61.31} & 36.05 & \underline{51.41} & \textbf{75.27} & \textbf{86.70} & {55.48} & {66.10} & \underline{51.61} & \underline{64.11} & \underline{81.63} & \underline{86.05} \\
\rowcolor{boarow}\textbf{\methodName (ours)} & \textbf{54.85} & \textbf{69.05} & \textbf{42.65} & {51.42} & \textbf{53.90} & \textbf{64.20} & \underline{36.65} & {50.83} & {61.70} & {78.32} & \underline{59.21} & \underline{68.31} & 51.49 & 63.69 & 81.37 & 85.27 \\
\midrule
\multicolumn{17}{c}{\textit{8$\times$~compression}} \\
\midrule
SAC & 49.60 & 62.12 & \underline{39.85} & \underline{49.37} & 42.70 & 53.03 & 32.05 & 47.35 & 46.54 & 61.47 & \textbf{60.61} & \textbf{69.22} & 45.23 & 57.09 & 68.10 & 72.96 \\
Beacon & \underline{53.66} & \underline{67.12} & 37.13 & 47.95 & \underline{48.23} & \underline{59.72} & \underline{33.68} & \textbf{50.08} & \textbf{75.27} & \textbf{86.29} & 51.29 & 62.25 & \textbf{49.88} & \underline{62.24} & \textbf{77.96} & \underline{82.56} \\
\rowcolor{boarow}\textbf{\methodName (ours)} & \textbf{57.18} & \textbf{71.18} & \textbf{40.45} & \textbf{50.22} & \textbf{52.50} & \textbf{62.52} & \textbf{34.12} & \underline{48.77} & \underline{55.70} & \underline{74.15} & \underline{58.55} & \underline{68.57} & \underline{49.75} & \textbf{62.57} & \underline{77.68} & \textbf{83.18} \\
\midrule
\multicolumn{17}{c}{\textit{16$\times$~compression}} \\
\midrule
SAC & 45.47 &	56.55&	31.00&	40.3&	34.24&	44.17&	25.37&	40.29&	38.26&	53.46&	52.95&	61.96&	37.88&	49.46&	52.50&	58.73 \\
Beacon & \underline{53.73} & \underline{65.22} & \underline{31.13} & \underline{41.47} & \underline{45.56} & \underline{56.44} & \underline{31.13} & \underline{47.03} & \textbf{74.35} & \textbf{85.69} & \underline{48.56} & \underline{58.91} & \underline{47.41} & \underline{59.13} & \underline{72.72} & \underline{76.76} \\
\rowcolor{boarow}\textbf{\methodName (ours)} & \textbf{54.39} & \textbf{68.91} & \textbf{43.31} & \textbf{52.02} & \textbf{52.10} & \textbf{62.56} & \textbf{34.57} & \textbf{48.48} & \underline{53.60} & \underline{71.62} & \textbf{58.55} & \textbf{67.84} & \textbf{49.42} & \textbf{61.91} & \textbf{76.98} & \textbf{81.95} \\
\bottomrule
\end{tabular}
\caption{MRQA \emph{out-of-domain} results with Qwen3-1.7B, per dataset. EM / F1; higher is better. \textbf{Norm~$\Delta$} normalises the average between the no-context floor (0) and full-context ceiling (100).  \methodName{} uses the Mamba-2 (1.3B) KV variant.}
\label{tab:ood}
\end{table*}

\subsection{Training Objective and Procedure}
\label{sec:method-train}

To align the representations between the Mamba encoder and the frozen Transformer decoder, \methodName employs a two-stage training strategy that updates only the encoder and projection layer.

\textbf{Stage 1: Pretraining.}
We pretrain the compressor on 10B tokens sampled from FineWeb-Edu \cite{penedo2024the} using a standard next-token cross-entropy objective, with a max sequence length of 1k tokens, conditioning the frozen decoder on the compressed sequence $E$. Causal language modelling provides a computationally lightweight base that transfers more effectively to compression than autoencoding \cite{liu2026sac}.
For our experiments in long-context encoding, we similarly pretrain on 10B tokens but selectively sampling the longest sequences of this dataset, targeting up to 10k sequence length.

\textbf{Stage 2: Distillation Fine-Tuning.}
We fine-tune the compressor using knowledge distillation from the uncompressed context teacher \cite{hinton2015distilling} across 500k QA pairs from the MRQA training dataset \cite{fisch-etal-2019-mrqa} for standard context compression, as well as dedicated long-context datasets (BookSum, GovReport, QASPER and LongAlpaca) \cite{kryscinski2021booksum, huang2021efficient, dasigi2021dataset, chen2023longlora} to evaluate long-context capabilities, training over samples of 10k sequence length. For each context, the compressor minimises the KL divergence over the answer token positions $t \in \{1, \dots, |y|\}$:

\begin{equation}
\label{eq:kl}
\mathcal{L} = \frac{\tau^2}{|y|}\sum_{t=1}^{|y|}\ KL\!\left( p_{\mathrm{T}}(\cdot \mid c, q, y_{<t}) \,\Vert\, p_{\mathrm{S}}(\cdot \mid Z, q, y_{<t}) \right) ,
\end{equation}

where $c$ is the uncompressed context, $q$ is the query, $y_{<t}$ are the previously generated answer tokens, $Z$ is the KV projection of the compressed state, $\tau$ is the distillation temperature, and $|y|$ is the total number of tokens in the generated answer sequence. Rather than forcing full token reconstruction, this objective directly aligns the compressed state with downstream generation by training the compressor to replicate full-context teacher behaviour \cite{cheng2024xrag}. All baselines share the same pretraining budget and data, with early stopping applied during fine-tuning. Further implementation and training details are provided in the Appendix.
\section{Experiments}
\label{sec:exp}

\subsection{Setup}
\label{sec:exp-setup}

\textbf{Benchmark.} We evaluate on MRQA, a suite of extractive question-answering datasets (divided into in-domain and out-of-domain tasks, where out-of-domain tasks are seen only at validation time). We report Exact Match (EM) and token-level F1 as fidelity metrics, averaged across tasks, and additionally a normalised score Norm$\Delta$ that places each configuration on a 0-100 scale between a no-context floor and a full-context ceiling, in order to see the true delta of the compressed context relative to the uncompressed baseline.

\textbf{Models.} We use two frozen decoders, Qwen3-1.7B and Qwen3-4B. For the compressor, we utilise pretrained Mamba-2 models at two scales: 1.3B and 2.7B. We evaluate compression ratios of 4$\times$, 8$\times$, and 16$\times$ with the KV-injection conditioning of Section~\ref{sec:method-cond}. We selected these specific power-of-two ratios because they are structurally required by the Activation Beacon baseline; the other evaluated methods are flexible enough to accommodate them, allowing for a fair and direct comparison across all approaches.

\textbf{Reference points.} We report compression performance relative to an upper and lower bound with respect to our target LLM model: Qwen3 fine-tuned using the complete context, which serves as the upper bound and is normalised to 100, and the same model evaluated without context, which serves as the lower bound and is normalised to 0. Based on our benchmark results across the evaluated state-of-the-art baselines, we select \textbf{Activation Beacon} as our primary compression reference as it achieves the strongest overall performance (different to the results reported in \citealt{liu2026sac} due to the use of compression ratios not supported by the Activation Beacon architecture).
We additionally compare against \textbf{SAC} to provide a comprehensive comparison with relevant methods in the context-compression setting. For methods not originally designed for this architecture, we adapted their implementations to Qwen3 to ensure a fair comparison.

We do not compare against RCC despite its SSM-based compressor architecture, as it fine-tunes the decoder and encoder jointly instead of conditioning a frozen LLM, and so does not provide a matched comparison.
In our benchmark, we trained each method to align with the data distribution of the target tasks. For Activation Beacon, we adapted the beacon window to accommodate the varying context lengths in MRQA. Although this configuration differs from that used in the original work, we preserved the same average number of beacon tokens per context.
Unless otherwise stated, all methods are evaluated in the online setting. Additional details regarding the experimental setup and hyperparameter configurations are presented in the Appendix.

\subsection{Compression Fidelity}
\label{sec:exp-quality}

The spread between the full-context ceiling and the no-context floor in Table~\ref{tab:indomain} confirms that the task chiefly depends on the context rather than on knowledge stored in model parameters. In-domain, \methodName scores above both baselines on both metrics at every ratio tested, recovering 95-96\% of the floor-to-ceiling F1 range against 72-88\% for Activation Beacon and 66-88\% for SAC. Out-of-domain (Table~\ref{tab:ood}) the three methods are much closer: SAC is strongest at 4$\times$, \methodName and Beacon are tied at 8$\times$, and \methodName dominates at 16$\times$, showing less degradation as compression ratio increases. The exception is Relation Extraction (RE), a task that requires verbatim extraction of text spans. This is the setting in which attention-based Transformers exceed SSMs, consistent with prior work \cite{jelassi2024repeat}, and here Beacon leads by 13.6--20.8 EM at every ratio.

\subsection{Compression Ratio Scaling}
\label{sec:exp-degradation}

While peak quality is similar across all methods at low compression ratios (4$\times$), \methodName degrades much more gracefully as compression ratio increases, losing less than 2\% of its in-domain EM, against roughly 10\% for Activation Beacon and 27\% for SAC, with the same ordering out-of-domain (Tables~\ref{tab:indomain} and~\ref{tab:ood}).
\methodName at 16$\times$ scores above Beacon at 4$\times$ in-domain, i.e.\ a four-fold shorter context at no cost in fidelity.
We refer to the Appendix for detailed visualisations of score degradation across compression ratios.

\begin{table}[!t]
\centering
\footnotesize
\setlength{\tabcolsep}{3.2pt}
\renewcommand{\arraystretch}{1.2}
\begin{tabular}{l cc cc cc}
\toprule
& \multicolumn{2}{c}{In-Domain} & \multicolumn{2}{c}{Out-of-Domain} & \multicolumn{2}{c}{Norm~$\Delta$} \\
\cmidrule(lr){2-3}\cmidrule(lr){4-5}\cmidrule(lr){6-7}
Config & EM & F1 & EM & F1 & EM & F1 \\
\midrule
\rowcolor{black!5}\textit{Full context (4B)} & 74.48 & 85.05 & 64.73 & 75.72 & \textcolor{green!60!black}{100} & \textcolor{green!60!black}{100} \\
\rowcolor{black!5}\textit{No context (4B)} & 15.39 & 20.60 & 18.17 & 23.76 & \textcolor{red!70!black}{0} & \textcolor{red!70!black}{0} \\
\midrule
\multicolumn{7}{c}{\textit{4$\times$~compression}} \\
\midrule
Beacon (4B) & 66.01 & 78.65 & \textbf{57.53} & \textbf{69.38} & 85.17 & \underline{89.06} \\
\rowcolor{boarow}\textbf{\methodName (2.7B)} & \underline{69.96} & \underline{81.65} & \underline{54.91} & \underline{66.88} & \textbf{86.43} & \textbf{89.49} \\
\rowcolor{boarow}\textbf{\methodName (1.3B)} & \textbf{70.63} & \textbf{81.88} & 53.67 & 65.30 & \underline{85.89} & 88.33 \\
\midrule
\multicolumn{7}{c}{\textit{8$\times$~compression}} \\
\midrule
Beacon (4B) & 64.08 & 76.29 & \textbf{56.71} & \textbf{68.25} & 82.57 & 86.06 \\
\rowcolor{boarow}\textbf{\methodName (2.7B)} & \textbf{70.14} & \textbf{81.96} & \underline{54.79} & \underline{67.14} & \textbf{86.48} & \textbf{89.98} \\
\rowcolor{boarow}\textbf{\methodName (1.3B)} & \underline{69.91} & \underline{81.76} & 53.57 & 65.72 & \underline{85.11} & \underline{88.58} \\
\midrule
\multicolumn{7}{c}{\textit{16$\times$~compression}} \\
\midrule
Beacon (4B) & 61.25 & 74.99 & \textbf{54.36} & \textbf{66.17} & 77.66 & 83.15 \\
\rowcolor{boarow}\textbf{\methodName (2.7B)} & \underline{66.83} & \underline{79.64} & \underline{52.48} & \underline{64.48} & \textbf{81.16} & \textbf{85.70} \\
\rowcolor{boarow}\textbf{\methodName (1.3B)} & \textbf{67.40} & \textbf{79.76} & 50.55 & 62.84 & \underline{79.88} & \underline{84.39} \\
\bottomrule
\end{tabular}
\caption{\textbf{Scalability to a larger decoder.} \methodName{} vs.\ Activation Beacon with a frozen \textbf{Qwen3-4B} decoder. \textbf{Norm~$\Delta$} pools the in- and out-of-domain averages and normalises between the pooled no-context floor (0) and full-context ceiling (100).}
\label{tab:scale-4b}
\end{table}

\subsection{Inference Latency}
\label{sec:exp-efficiency}

\begin{figure}[!t]
\centering
\includegraphics[width=0.95\columnwidth]{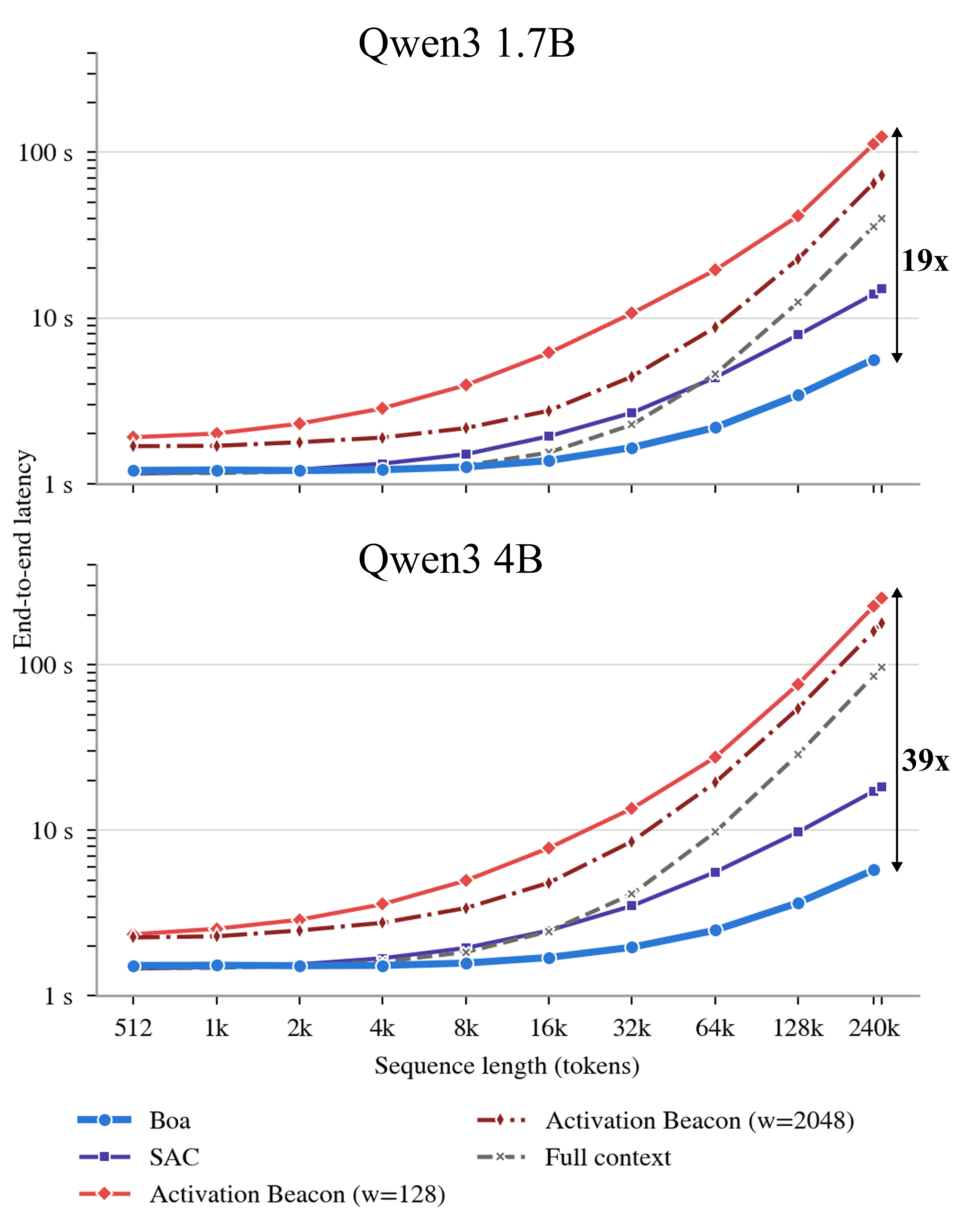}
\caption{Total processing time versus input length at 8$\times$ compression: \methodName against other methods and full-context processing, up to 240K tokens for Qwen3-1.7B and 4B. We included both windows sizes \emph{w} used for short context and long context for Beacon}
\label{fig:beacon}
\end{figure}

Figure~\ref{fig:beacon} reports end-to-end latency (encoding, prefill, and generation) against input length using a frozen Qwen3 backbone (comparing the 1.7B and 4B Qwen checkpoints).
The baseline curve is full-context processing by the unmodified LLM. The other curves report inference latency for each comparison method in the online compression regime (with single-use context). \methodName operates consistently at or below this baseline, achieving a net speedup without requiring context re-use as in the offline compression case.
We note that SAC achieves moderate speedups at longer contexts, while Activation Beacon carries a significant inference cost that scales quadratically with prefill length, never reaching viability for online use.
\methodName maintains lower compute latency across all evaluated sequence lengths and decoder sizes, while matching or outperforming comparisons in performance at almost every compression ratio.
A more detailed analysis of computational cost, memory overhead and latency is presented in the Appendix.

\subsection{Long-Context Evaluation}

Table~\ref{tab:long_ctx} reports results on long context benchmarks (up to 32K tokens). On summarisation tasks BookSum and GovReport, \methodName achieves performance comparable to Activation Beacon. However, Activation Beacon retains an advantage on the question-answering task QASPER. We attribute this gap to a weaker long-context prior, as the baseline Mamba checkpoint was pretrained solely on sequence lengths up to 2K tokens, whereas Activation Beacon operates directly on full-attention windows across extended contexts.

\subsection{Encoder-Decoder Scale Ratios}
\label{sec:exp-scale}

Table~\ref{tab:scale-4b} evaluates \methodName across two compressor scales (1.3B and 2.7B parameters) using a frozen Qwen3-4B decoder.
Note that while Activation Beacon inherently uses the full 4B Qwen backbone for both encoding and decoding, \methodName achieves comparable performance with significantly smaller, lightweight encoders, even when the LLM decoder is scaled up to compensate.
Doubling our compressor to 2.7B leaves in-domain accuracy unchanged ($\le$0.7 EM) and yields only small out-of-domain gains (1.2--1.9 EM).
Extended results and full tables are provided in the Appendix.

\subsection{Ablations}
\label{sec:exp-ablation}

Table~\ref{tab:ablation-main} separates the architectural choices from the training objective. The choice of conditioning space is an order of magnitude more impactful than the choice of encoder family, which is consistent with 500xCompressor~\cite{li2024500x}.
As a training objective, distillation is substantially more effective in the fine-tuning stage than as a off-policy pretraining signal: moving it to pretraining recovers less than half of the in-domain difference between the full recipe and language modelling alone.

Table~\ref{tab:ablation-summary} shows that the three components act on different axes and are not interchangeable: distillation fine-tuning contributes strongly to in-domain performance, KV injection makes the largest difference for out-of-domain performance, and language-modelling pretraining contributes moderately to both. Each removal falls below the Activation Beacon baseline on at least one split.

\begin{table}[!t]
\centering
\footnotesize
\setlength{\tabcolsep}{6pt}
\renewcommand{\arraystretch}{1.2}
\begin{tabular}{l c c c | c }
\toprule
Method & BookSum & GovReport & Qasper & Average \\
\midrule
\rowcolor{black!5}\textit{Full context} & 18.84 & 30.17 & 37.11 & 28.71 \\
\rowcolor{black!5}\textit{No context} & 4.89 & 12.21 & 11.06 & 9.39 \\
\midrule
\multicolumn{5}{c}{\textit{8$\times$~compression}} \\
\midrule
Beacon & \textbf{17.79} & 24.17 & \textbf{32.24} & \textbf{24.73} \\
\rowcolor{boarow}\textbf{Boa (ours)} & 17.01 & \textbf{26.07} & 26.82 & 23.30 \\
\midrule
\multicolumn{5}{c}{\textit{16$\times$~compression}} \\
\midrule
Beacon & \textbf{18.00} & 21.18 & \textbf{30.85} & \textbf{23.34} \\
\rowcolor{boarow}\textbf{Boa (ours)} & 16.59 & \textbf{25.13} & 24.50 & 22.07 \\
\bottomrule
\end{tabular}
\caption{\textbf{Long-context} results at 32K sequence length under 8$\times$ and 16$\times$ compression ratios against Activation Beacon.}
\label{tab:long_ctx}
\end{table}

\begin{table}[!t]
\centering
\small
\setlength{\tabcolsep}{4pt}
\begin{tabular}{l cc cc}
\toprule
& \multicolumn{2}{c}{In-Domain} & \multicolumn{2}{c}{Out-of-Domain} \\
\cmidrule(lr){2-3}\cmidrule(lr){4-5}
Config & EM & F1 & EM & F1 \\
\midrule
\multicolumn{5}{c}{\textit{Encoder $\times$ projection}} \\
\midrule
Mamba-1, Emb & 64.90 & 76.37 & 42.70 & 55.01 \\
Mamba-1, KV & \underline{68.88} & \underline{80.09} & \underline{49.00} & \underline{61.21} \\
Mamba-2, Emb & 64.48 & 76.19 & 42.92 & 55.40 \\
\rowcolor{boarow}\textbf{Mamba-2, KV \;(=\,\methodName)} & \textbf{69.02} & \textbf{80.46} & \textbf{49.75} & \textbf{62.57} \\
\midrule
\multicolumn{5}{c}{\textit{Training objective: pretrain $\rightarrow$ finetune \;(Mamba-2, KV)}} \\
\midrule
\rowcolor{boarow}\textbf{LM $\rightarrow$ KD \;(=\,\methodName)} & \textbf{69.02} & \textbf{80.46} & \textbf{49.75} & \textbf{62.57} \\
LM $\rightarrow$ LM & 63.15 & 76.29 & \underline{49.08} & 61.00 \\
Off-policy KD $\rightarrow$ LM & \underline{65.03} & \underline{77.92} & \textbf{49.75} & \underline{62.25} \\
\bottomrule
\end{tabular}
\caption{\textbf{Ablations} at $8\times$ (Qwen3-1.7B). Top: encoder $\times$ conditioning space. Bottom: training objective.}
\label{tab:ablation-main}
\end{table}

\section{Discussion}
\label{sec:disc}

Our experiments in Section~\ref{sec:exp} show that \methodName matches state-of-the-art compression performance across a range of benchmarks while substantially reducing inference time. These results, along with our ablations, demonstrate that KV cache injection works well for compression in spite of an architectural mismatch between the frozen LLM decoder and a trained, separate encoder. This suggests that small, efficient encoders can be effectively paired with larger reasoning decoders, avoiding the LLM prefill bottleneck and enabling online compression even for very large Transformer models.

Our approach presents clear scaling potential for future research. \methodName achieves competitive fidelity despite a significant pretraining deficit relative to existing methods. While techniques that reuse the target LLM benefit from representations honed over trillions of tokens and extended context windows, our Mamba encoders were trained on approximately $300\text{B}$ tokens and exposed only to $2\text{K}$-token sequences. The observation that our approach matches or surpasses state-of-the-art compressors under these constraints suggests that the current results serve as a baseline for future optimisation.

Several directions for future work naturally follow from this foundation. First, initialising the compressor with state-space models trained on larger token budgets and natively longer sequences may address current capacity limitations and further improve compression performance. Additionally, the Boa architecture could be extended to support adaptive compression of varying token slices. Rather than relying on a fixed compression ratio with a static strided slice parameter, future iterations could dynamically adjust the compression rate based on the semantic complexity of the context.
\begin{table}[!t]
\centering
\small
\setlength{\tabcolsep}{4pt}
\begin{tabular}{l cc cc}
\toprule
& \multicolumn{2}{c}{In-Domain} & \multicolumn{2}{c}{Out-of-Domain} \\
\cmidrule(lr){2-3}\cmidrule(lr){4-5}
Config & EM & F1 & EM & F1 \\
\midrule
\rowcolor{boarow}\textbf{\methodName{} (full)} & \textbf{69.02} & \textbf{80.46} & \textbf{49.75} & \textbf{62.57} \\
\quad w/o LM pretraining & \underline{66.55} & \underline{78.97} & 47.22 & 60.04 \\
\quad w/o KD finetuning & 63.15 & 76.29 & \underline{49.08} & \underline{61.00} \\
\quad w/o KV projection \;(Emb) & 64.48 & 76.19 & 42.92 & 55.40 \\
\bottomrule
\end{tabular}
\caption{\textbf{Component ablation} of \methodName{} at $8\times$. Each row removes one component from the full model.}
\label{tab:ablation-summary}
\end{table}

\section{Conclusion}
\label{sec:conc}

We presented \methodName, a context compressor that uses a linear-time state-space model to encode arbitrary input context into a representation that conditions a frozen Transformer decoder, trained through knowledge distillation, with no modification or fine-tuning of the decoder. \methodName matches or surpasses a strong compression baseline at matched compression ratios, degrades far more gracefully as the ratio increases, and reduces inference latency and memory by large margins. This makes compression practically viable online at inference time rather than only as offline preprocessing.

\clearpage
\bibliographystyle{plainnat}
\bibliography{main}

\clearpage
\onecolumn
\appendix

\section{Implementation and Training Details}

\subsection{MRQA Training}
Table~\ref{tab:training_config} reports the full configuration of the two
training stages of the MRQA
\methodName{} pipeline: (i) \emph{pretraining} of the Mamba2
compressor with a next-token language-modeling objective, and (ii)
\emph{knowledge-distillation (KD) fine-tuning} on MRQA against the frozen
teacher. In both stages the Qwen3-1.7B decoder is kept frozen and the
compressed context is injected through the decoder's KV cache. All hyperparameters below are shared across \emph{every} compression ratio ($4\times$, $8\times$, $16\times$); the ratio is
controlled by a single \texttt{stride} argument and is the only value that
changes between runs. The fine-tuning stage is initialized from the pretrained
compressor checkpoint of the matching ratio.

\paragraph{Frozen teacher: Qwen3-1.7B full fine-tuning on MRQA.}
The frozen decoder used throughout is Qwen3-1.7B, obtained by fully
fine-tuning (FullFT) the base Qwen3-1.7B checkpoint on MRQA. We train for a
single epoch, which we find sufficient for convergence on this dataset. This
FullFT checkpoint is the teacher/decoder used by every \methodName{}
pretraining and fine-tuning run reported here. Table~\ref{tab:fullft_config}
summarizes its configuration.

\begin{table}[htbp]
\centering
\small
\setlength{\tabcolsep}{5pt}
\renewcommand{\arraystretch}{1.15}
\begin{tabular}{@{}l l@{}}
\toprule
\textbf{Setting} & \textbf{Value} \\
\midrule
Base model            & Qwen3-1.7B \\
Dataset               & MRQA \\
Epochs                & $1$ \\
Learning rate         & $4\times10^{-5}$ \\
LR schedule           & cosine w/ min LR (min ratio $0.1$) \\
Warmup steps          & $100$ \\
Per-device batch      & $16$ \\
Grad accumulation     & $1$ \\
Optimizer             & AdamW (fused), wd $0.01$ \\
Grad clip (max norm)  & $1.0$ \\
Precision             & bfloat16, grad.\ checkpointing \\
Seed                  & $42$ \\
\bottomrule
\end{tabular}
\caption{Full fine-tuning configuration of the Qwen3-1.7B teacher on MRQA.}
\label{tab:fullft_config}
\end{table}

\subsection{Long-Context Training}

\begin{table}[t]
\centering
\small
\setlength{\tabcolsep}{6pt}
\renewcommand{\arraystretch}{1.2}
\begin{tabular}{@{}l p{6.1cm} p{6.1cm}@{}}
\toprule
\textbf{Setting} & \textbf{Pretraining} & \textbf{KD fine-tuning} \\
\midrule
\multicolumn{3}{@{}l}{\emph{Model / architecture}} \\
\quad Compressor            & Mamba2-1.3B (\texttt{mamba2-1.3b}) & Mamba2-1.3B (\texttt{mamba2-1.3b}) \\
\quad Compressor tokenizer  & EleutherAI/gpt-neox-20b & EleutherAI/gpt-neox-20b \\
\quad Decoder (frozen)      & Qwen3-1.7B & Qwen3-1.7B \\
\quad Context injection     & KV cache (\texttt{kv}) & KV cache (\texttt{kv}) \\
\quad Initialization        & Mamba2-1.3B pretrained weights & pretrained compressor (matching ratio) \\
\midrule
\multicolumn{3}{@{}l}{\emph{Objective / data}} \\
\quad Loss                  & LM continuation (cross-entropy) & KL distillation (\texttt{kl}) \\
\quad Distillation temp.    & -- & $2.0$ \\
\quad Corpus / dataset      & FineWeb-Edu (10B-token sample) & MRQA \\
\quad Context length        & $1024$ (\texttt{max\_ctx}) & $1024$ (\texttt{max\_seq\_len}) \\
\quad Prediction length     & $256$ (\texttt{max\_pred}) & $80$ (\texttt{max\_new\_tokens}) \\
\quad Min.\ doc.\ tokens     & $320$ & -- \\
\midrule
\multicolumn{3}{@{}l}{\emph{Optimization}} \\
\quad Optimizer             & AdamW & AdamW \\
\quad Learning rate         & $1.0\times10^{-4}$ & $1.0\times10^{-4}$ \\
\quad Min.\ LR ($\eta_{\min}$) & -- & $1.0\times10^{-6}$ \\
\quad Weight decay          & $0.01$ & $0.01$ \\
\quad LR schedule           & cosine w/ warmup & cosine w/ warmup \\
\quad Warmup steps          & $50$ & $400$ \\
\quad Grad clip (max norm)  & $1.0$ & $2.0$ \\
\quad Duration              & until $10$B tokens & early stopping (patience $6$) \\
\midrule
\multicolumn{3}{@{}l}{\emph{Batching / runtime}} \\
\quad Packing               & token-budget ($4096$ tok/batch, $\le 32$ seqs) & none (plain batches) \\
\quad Per-GPU batch         & $8$ & $4$ \\
\quad Grad accumulation     & $4$ & $2$ \\
\quad Parallelism           & DDP, $32$ nodes ($128$ GPUs) & DDP, $32$ nodes ($128$ GPUs) \\
\quad Decoder dtype         & bfloat16 & bfloat16 \\
\quad Seed                  & $42$ & $42$ \\
\bottomrule
\end{tabular}
\caption{Training configuration of the two \methodName{} stages on MRQA:
Language Modeling (LM) pretraining of the Mamba2 compressor (left) and KL
knowledge-distillation fine-tuning against the frozen Qwen3-1.7B teacher
(right). All values are shared across every compression ratio
($4\times/8\times/16\times$); only \texttt{stride} (and, for fine-tuning, the
matching pretrained init checkpoint) changes between runs.}
\label{tab:training_config}
\end{table}

The initial setup used for the long-context training is the same as specified
above, with the same model configurations. The training setup is changed in
order to increase the sequence length employed, mainly adding full activation
checkpointing, a float8 variant of AdamW, and making use of common PyTorch
optimisation techniques; Table~\ref{tab:longctx_training_config} reports the
resulting configuration.

\paragraph{Long-context pretraining data.}
We sample the longest sequences from the FineWeb-Edu 100B pool of data. Since
we truncate the context during training if it surpasses our chosen sequence
length, we sampled more than 10B tokens from the 100B pool in order to reach
our 10B-token training budget target.

\begin{figure}[!t]
\centering
\includegraphics[width=0.8\linewidth]{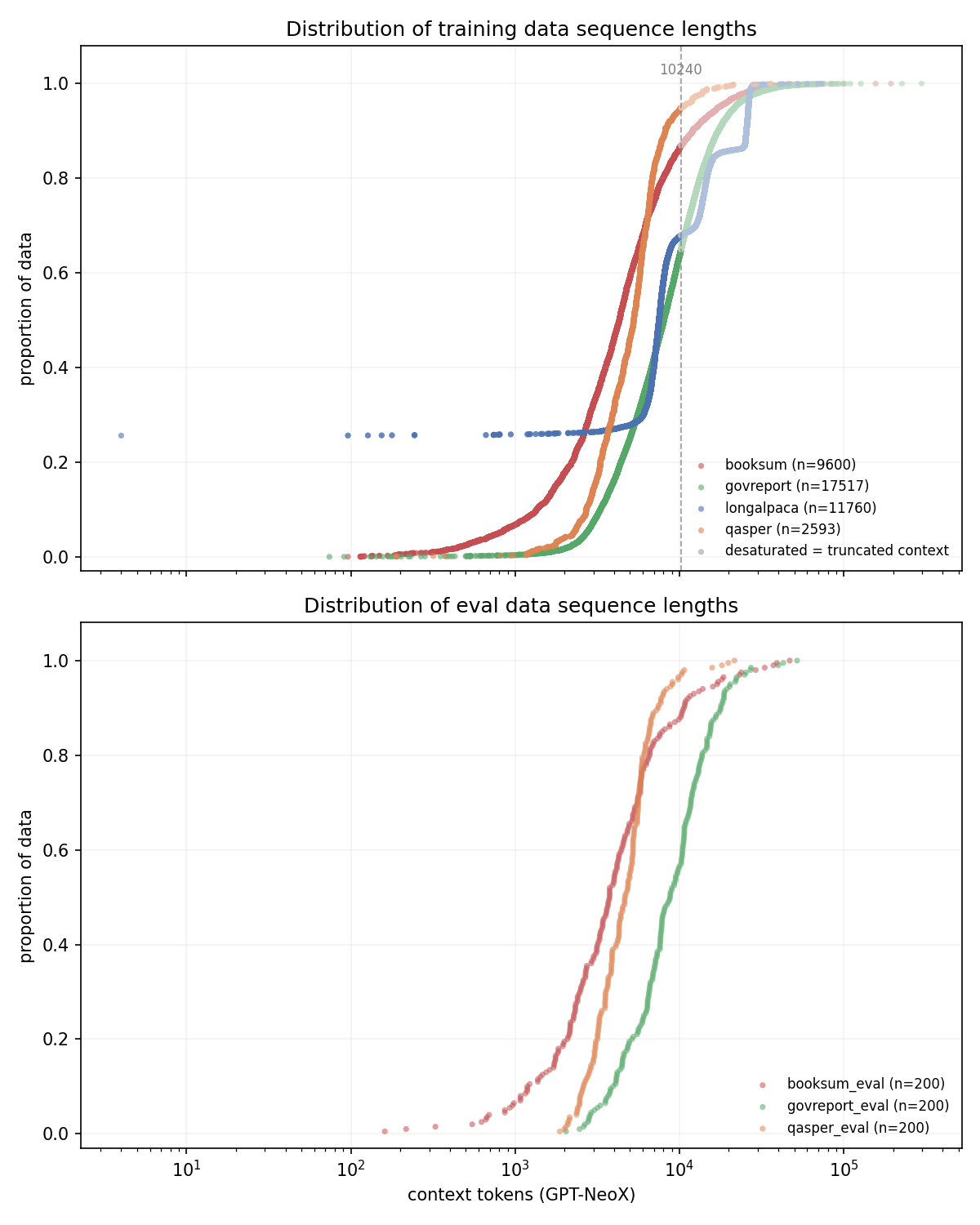}
\caption{Sequence-length distributions of the long-context KD fine-tuning data,
shown as empirical cumulative distributions (each point is one sample; the
$y$-axis gives the proportion of samples at or below its length, the $x$-axis
the context length in \methodName{} tokens, log-scaled).
\textbf{Top:} training corpora (LongAlpaca-12k, BookSum, GovReport, Qasper);
the dashed line marks the $10{,}240$-token training sequence length, and
samples beyond it (drawn desaturated) are truncated during training.
\textbf{Bottom:} evaluation splits. GovReport is the longest on average and
has the most samples above the training sequence length. We evaluate on the
full sequence length of each sample without truncation.}
\label{fig:longctx_seqlen}
\end{figure}

\paragraph{Long-context fine-tuning data.}
The following datasets are used for the fine-tuning phase using KD in the
long-context setups: LongAlpaca-12k, BookSum, GovReport, and Qasper. BookSum
and GovReport correspond to summarisation tasks, while Qasper and LongAlpaca
correspond to question answering. Figure~\ref{fig:longctx_seqlen} provides the
sequence-length distribution of the data used for evaluation and training
during this phase. During training, at $10{,}240$ tokens we start truncating the
context of the given sample. Token counts are calculated using the tokeniser
employed by \methodName{}.

\begin{table}[!tb]
\centering
\footnotesize
\setlength{\tabcolsep}{4pt}
\renewcommand{\arraystretch}{1.02}
\begin{tabular}{@{}l p{5.4cm} p{6.4cm}@{}}
\toprule
\textbf{Setting} & \textbf{Long-context pretraining} & \textbf{KD fine-tuning} \\
\midrule
\multicolumn{3}{@{}l}{\emph{Model / architecture}} \\
\quad Compressor            & Mamba2-1.3B                  & Mamba2-1.3B \\
\quad Compressor tokenizer  & EleutherAI/gpt-neox-20b      & EleutherAI/gpt-neox-20b \\
\quad Decoder (frozen)      & Qwen3-1.7B                   & Qwen3-1.7B \\
\quad Context injection     & KV cache (\texttt{kv})       & KV cache (\texttt{kv}) \\
\quad Initialization        & Mamba2-1.3B pretrained weights & pretrained compressor (matching ratio) \\
\midrule
\multicolumn{3}{@{}l}{\emph{Objective}} \\
\quad Loss                  & LM continuation (cross-entropy) & KL distillation (\texttt{kl}) \\
\quad Max new / pred tokens & $256$ (\texttt{max\_pred})    & $256$ (\texttt{max\_new\_tokens}) \\
\midrule
\multicolumn{3}{@{}l}{\emph{Data}} \\
\quad Corpus                & FineWeb-Edu (10B longest samples) & Long-context mixture (LongAlpaca, BookSum, GovReport, Qasper) \\
\quad Distillation temp.    & --                           & $2.0$ \\
\quad Context length        & $10{,}240$ (\texttt{max\_ctx}) & $10{,}240$ (\texttt{max\_seq\_len}) \\
\midrule
\multicolumn{3}{@{}l}{\emph{Optimization}} \\
\quad Optimizer             & AdamW-float8          & AdamW-float8\\
\quad Learning rate         & $1.0\times10^{-4}$           & $1.0\times10^{-4}$ \\
\quad Min.\ LR ($\eta_{\min}$) & --                        & $1.0\times10^{-6}$ \\
\quad Weight decay          & $0.01$                       & $0.01$ \\
\quad LR schedule           & cosine w/ warmup             & cosine w/ warmup \\
\quad Warmup steps          & $50$                         & $400$ \\
\quad Grad clip (max norm)  & $1.0$                        & $2.0$ \\
\quad Duration              & until $10$B tokens           & early stopping (patience $4$) \\
\midrule
\multicolumn{3}{@{}l}{\emph{Batching / parallelism}} \\
\quad Per-GPU batch         & $1$ (token-budget packed)    & $1$ \\
\quad Packing               & token-budget                 & none (plain batches) \\
\quad Grad accumulation     & $2$                          & $2$ \\
\quad Parallelism           & DDP, $32$ nodes ($128$ GPUs) & DDP, $32$ nodes ($128$ GPUs) \\
\quad Decoder dtype         & bfloat16                     & bfloat16 \\
\quad Seed                  & $42$                         & $42$ \\
\bottomrule
\end{tabular}
\caption{Training configuration of the two \methodName{} stages on long-context
data: Language Modeling (LM) pretraining of the Mamba2 compressor (left) and KL
knowledge-distillation fine-tuning against the frozen Qwen3-1.7B teacher
(right). All values are shared for both $8\times$ and $16\times$ compression
ratios; only \texttt{stride} (and, for fine-tuning, the matching pretrained
init checkpoint) changes between runs.}
\label{tab:longctx_training_config}
\end{table}

\subsection{Hardware}
All runs were trained on the accelerated (ACC) partition of the MareNostrum~5
supercomputer using \textbf{32 nodes}. Each node is equipped with
\textbf{4$\times$ NVIDIA Hopper H100 with 64\,GB HBM2}, two Intel Xeon Platinum
8460Y+ (40C, 2.3\,GHz) CPUs, and 512\,GB of DDR5 memory, interconnected with
NDR200 InfiniBand. This yields a total of \textbf{128 H100 GPUs}. Training uses
distributed data parallelism (DDP) across all 128 GPUs, with the frozen decoder
run in bfloat16.

\section{Scalability to a Larger Decoder: Full Per-Dataset Results}
\label{sec:supp-scale4b}

The scalability table in the main paper reports in-domain and
out-of-domain averages for \methodName{} and Activation Beacon with a frozen
\textbf{Qwen3-4B} decoder, at compressor scales 1.3B and 2.7B.
Tables~\ref{tab:scale4b-indomain} and~\ref{tab:scale4b-ood} below give the
full per-dataset breakdown behind those averages.

\begin{table}[!t]
\centering
\footnotesize
\setlength{\tabcolsep}{3.2pt}
\renewcommand{\arraystretch}{1.2}
\resizebox{\textwidth}{!}{
\begin{tabular}{@{}l cc cc cc cc cc cc @{\hspace{6pt}} cc @{\hspace{6pt}} cc@{}}
\toprule
& \multicolumn{2}{c}{SQuAD} & \multicolumn{2}{c}{NewsQA} & \multicolumn{2}{c}{TriviaQA} & \multicolumn{2}{c}{SearchQA} & \multicolumn{2}{c}{HotpotQA} & \multicolumn{2}{c}{NaturalQ} & \multicolumn{2}{c}{Average} & \multicolumn{2}{c}{Norm~$\Delta$} \\
\cmidrule(lr){2-3}\cmidrule(lr){4-5}\cmidrule(lr){6-7}\cmidrule(lr){8-9}\cmidrule(lr){10-11}\cmidrule(lr){12-13}\cmidrule(lr){14-15}\cmidrule(lr){16-17}
Method & EM & F1 & EM & F1 & EM & F1 & EM & F1 & EM & F1 & EM & F1 & EM & F1 & EM & F1 \\
\midrule
\rowcolor{black!5}\textit{Full context (ceiling)} & 89.42 & 94.98 & 59.66 & 74.15 & 76.87 & 84.87 & 76.84 & 86.62 & 70.89 & 85.46 & 73.22 & 84.20 & 74.48 & 85.05 & \textcolor{green!60!black}{100} & \textcolor{green!60!black}{100} \\
\rowcolor{black!5}\textit{No context (floor)} & 8.86 & 16.61 & 2.40 & 5.34 & 35.57 & 39.16 & 28.51 & 32.28 & 6.52 & 12.76 & 10.46 & 17.43 & 15.39 & 20.60 & \textcolor{red!70!black}{0} & \textcolor{red!70!black}{0} \\
\midrule
\multicolumn{17}{c}{\textit{4$\times$~compression}} \\
\midrule
Beacon (4B) & 70.83 & 84.83 & 52.42 & 69.79 & 71.33 & 78.33 & 69.68 & 77.62 & \textbf{66.48} & \textbf{82.06} & 65.29 & 79.24 & 66.00 & 78.64 & 85.66 & 90.06 \\
\rowcolor{boarow}\textbf{\methodName (2.7B)} & \underline{80.90} & \underline{90.38} & \underline{53.47} & \underline{69.92} & \underline{75.48} & \underline{82.66} & \underline{78.69} & \underline{86.48} & 64.29 & 80.67 & \textbf{66.91} & \textbf{79.82} & \underline{69.96} & \underline{81.65} & \underline{92.34} & \underline{94.74} \\
\rowcolor{boarow}\textbf{\methodName (1.3B)} & \textbf{81.48} & \textbf{90.41} & \textbf{54.34} & \textbf{70.49} & \textbf{76.69} & \textbf{83.33} & \textbf{79.45} & \textbf{86.69} & \underline{65.09} & \underline{80.78} & \underline{66.74} & \underline{79.60} & \textbf{70.63} & \textbf{81.88} & \textbf{93.49} & \textbf{95.09} \\
\midrule
\multicolumn{17}{c}{\textit{8$\times$~compression}} \\
\midrule
Beacon (4B) & 68.46 & 83.56 & 50.07 & 68.30 & 70.49 & 77.52 & 68.35 & 70.40 & 64.12 & 80.25 & 62.99 & 77.71 & 64.08 & 76.29 & 82.40 & 86.41 \\
\rowcolor{boarow}\textbf{\methodName (2.7B)} & \textbf{81.12} & \textbf{90.48} & \textbf{53.63} & \underline{70.01} & \textbf{75.71} & \textbf{83.51} & \underline{78.81} & \underline{86.89} & \textbf{64.87} & \underline{80.76} & \textbf{66.73} & \textbf{80.08} & \textbf{70.14} & \textbf{81.96} & \textbf{92.66} & \textbf{95.20} \\
\rowcolor{boarow}\textbf{\methodName (1.3B)} & \underline{79.73} & \underline{89.75} & \underline{53.18} & \textbf{70.10} & \underline{75.30} & \underline{82.69} & \textbf{80.47} & \textbf{87.43} & \underline{64.60} & \textbf{80.85} & \underline{66.17} & \underline{79.75} & \underline{69.91} & \underline{81.76} & \underline{92.26} & \underline{94.90} \\
\midrule
\multicolumn{17}{c}{\textit{16$\times$~compression}} \\
\midrule
Beacon (4B) & 65.43 & 81.16 & 47.15 & 66.21 & 68.46 & 75.76 & 65.57 & 73.69 & 61.50 & 78.36 & 59.40 & 74.73 & 61.25 & 74.98 & 77.61 & 84.38 \\
\rowcolor{boarow}\textbf{\methodName (2.7B)} & \textbf{77.05} & \textbf{87.98} & \textbf{50.95} & \textbf{68.21} & \underline{73.19} & \underline{81.31} & \underline{74.51} & \underline{84.04} & \underline{61.80} & \underline{78.65} & \textbf{63.45} & \textbf{77.63} & \underline{66.83} & \underline{79.64} & \underline{87.05} & \underline{91.61} \\
\rowcolor{boarow}\textbf{\methodName (1.3B)} & \underline{76.90} & \underline{87.40} & \underline{49.50} & \underline{67.60} & \textbf{75.16} & \textbf{82.17} & \textbf{78.20} & \textbf{85.81} & \textbf{61.96} & \textbf{78.76} & \underline{62.68} & \underline{76.83} & \textbf{67.40} & \textbf{79.76} & \textbf{88.01} & \textbf{91.80} \\
\bottomrule
\end{tabular}
}
\caption{MRQA \emph{in-domain} results with a frozen \textbf{Qwen3-4B} decoder, per dataset. EM / token-level F1; higher is better. \textbf{Norm~$\Delta$} normalises the average between the no-context floor (0) and full-context ceiling (100).}
\label{tab:scale4b-indomain}
\end{table}

\begin{table}[!t]
\centering
\footnotesize
\setlength{\tabcolsep}{3.2pt}
\renewcommand{\arraystretch}{1.2}
\resizebox{\textwidth}{!}{
\begin{tabular}{@{}l cc cc cc cc cc cc @{\hspace{6pt}} cc @{\hspace{6pt}} cc@{}}
\toprule
& \multicolumn{2}{c}{BioASQ} & \multicolumn{2}{c}{DROP} & \multicolumn{2}{c}{DuoRC} & \multicolumn{2}{c}{RACE} & \multicolumn{2}{c}{RE} & \multicolumn{2}{c}{TextbookQA} & \multicolumn{2}{c}{Average} & \multicolumn{2}{c}{Norm~$\Delta$} \\
\cmidrule(lr){2-3}\cmidrule(lr){4-5}\cmidrule(lr){6-7}\cmidrule(lr){8-9}\cmidrule(lr){10-11}\cmidrule(lr){12-13}\cmidrule(lr){14-15}\cmidrule(lr){16-17}
Method & EM & F1 & EM & F1 & EM & F1 & EM & F1 & EM & F1 & EM & F1 & EM & F1 & EM & F1 \\
\midrule
\rowcolor{black!5}\textit{Full context (ceiling)} & 66.82 & 81.44 & 63.81 & 72.15 & 60.23 & 70.89 & 51.93 & 65.22 & 79.85 & 89.51 & 65.74 & 75.13 & 64.73 & 75.72 & \textcolor{green!60!black}{100} & \textcolor{green!60!black}{100} \\
\rowcolor{black!5}\textit{No context (floor)} & 30.12 & 37.49 & 22.22 & 26.04 & 2.80 & 5.74 & 2.52 & 7.05 & 12.62 & 19.45 & 38.72 & 46.78 & 18.17 & 23.76 & \textcolor{red!70!black}{0} & \textcolor{red!70!black}{0} \\
\midrule
\multicolumn{17}{c}{\textit{4$\times$~compression}} \\
\midrule
Beacon (4B) & \textbf{58.18} & \textbf{72.44} & \textbf{53.83} & \textbf{63.39} & 53.63 & \underline{65.15} & \underline{39.61} & \textbf{55.64} & \textbf{79.58} & \textbf{89.05} & 60.35 & \textbf{70.54} & \textbf{57.53} & \textbf{69.37} & \textbf{84.53} & \textbf{87.78} \\
\rowcolor{boarow}\textbf{\methodName (2.7B)} & \underline{58.05} & \underline{71.82} & \underline{49.90} & \underline{58.12} & \underline{54.36} & 64.83 & \textbf{40.50} & \underline{54.91} & 65.64 & \underline{81.21} & \underline{61.01} & \underline{70.38} & \underline{54.91} & \underline{66.88} & \underline{78.91} & \underline{82.98} \\
\rowcolor{boarow}\textbf{\methodName (1.3B)} & 56.72 & 70.27 & 45.44 & 53.87 & \textbf{55.43} & \textbf{65.24} & 37.09 & 51.25 & \underline{66.28} & 81.07 & \textbf{61.08} & 70.09 & 53.67 & 65.30 & 76.26 & 79.94 \\
\midrule
\multicolumn{17}{c}{\textit{8$\times$~compression}} \\
\midrule
Beacon (4B) & \underline{58.31} & 71.66 & \underline{49.63} & \textbf{59.46} & 53.30 & 64.52 & \textbf{39.17} & \textbf{54.68} & \textbf{78.56} & \textbf{88.46} & 61.34 & 70.72 & \textbf{56.72} & \textbf{68.25} & \textbf{82.80} & \textbf{85.62} \\
\rowcolor{boarow}\textbf{\methodName (2.7B)} & \textbf{58.91} & \textbf{73.48} & \textbf{50.30} & \underline{58.87} & \textbf{57.03} & \textbf{66.55} & \underline{39.02} & \underline{54.27} & 61.43 & \underline{78.34} & \underline{62.08} & \underline{71.35} & \underline{54.79} & \underline{67.14} & \underline{78.66} & \underline{83.49} \\
\rowcolor{boarow}\textbf{\methodName (1.3B)} & 58.11 & \underline{72.35} & 47.70 & 56.68 & \underline{54.63} & \underline{64.89} & 36.65 & 51.55 & \underline{61.50} & 77.33 & \textbf{62.81} & \textbf{71.56} & 53.57 & 65.72 & 76.03 & 80.76 \\
\midrule
\multicolumn{17}{c}{\textit{16$\times$~compression}} \\
\midrule
Beacon (4B) & \underline{56.58} & 69.70 & \textbf{45.44} & \textbf{55.38} & 51.10 & \underline{62.58} & \textbf{37.24} & \textbf{52.67} & \textbf{78.05} & \textbf{88.35} & 57.75 & 68.36 & \textbf{54.36} & \textbf{66.17} & \textbf{77.73} & \textbf{81.62} \\
\rowcolor{boarow}\textbf{\methodName (2.7B)} & \textbf{59.91} & \textbf{73.05} & \underline{45.04} & \underline{53.67} & \textbf{52.83} & \textbf{62.82} & \underline{36.35} & \underline{50.96} & \underline{59.53} & \underline{76.17} & \textbf{61.21} & \textbf{70.20} & \underline{52.48} & \underline{64.48} & \underline{73.69} & \underline{78.36} \\
\rowcolor{boarow}\textbf{\methodName (1.3B)} & 56.18 & \underline{69.84} & 44.51 & 53.05 & \underline{51.70} & 62.20 & 33.38 & 48.41 & 58.18 & 74.17 & \underline{59.35} & \underline{69.34} & 50.55 & 62.84 & 69.55 & 75.20 \\
\bottomrule
\end{tabular}
}
\caption{MRQA \emph{out-of-domain} results with a frozen \textbf{Qwen3-4B} decoder, per dataset. EM / F1; higher is better. \textbf{Norm~$\Delta$} normalises the average between the no-context floor (0) and full-context ceiling (100).}
\label{tab:scale4b-ood}
\end{table}

\section{Latency and Memory Efficiency}
\label{sec:supp-latency}

\begin{figure}[!t]
\centering
\includegraphics[width=\linewidth]{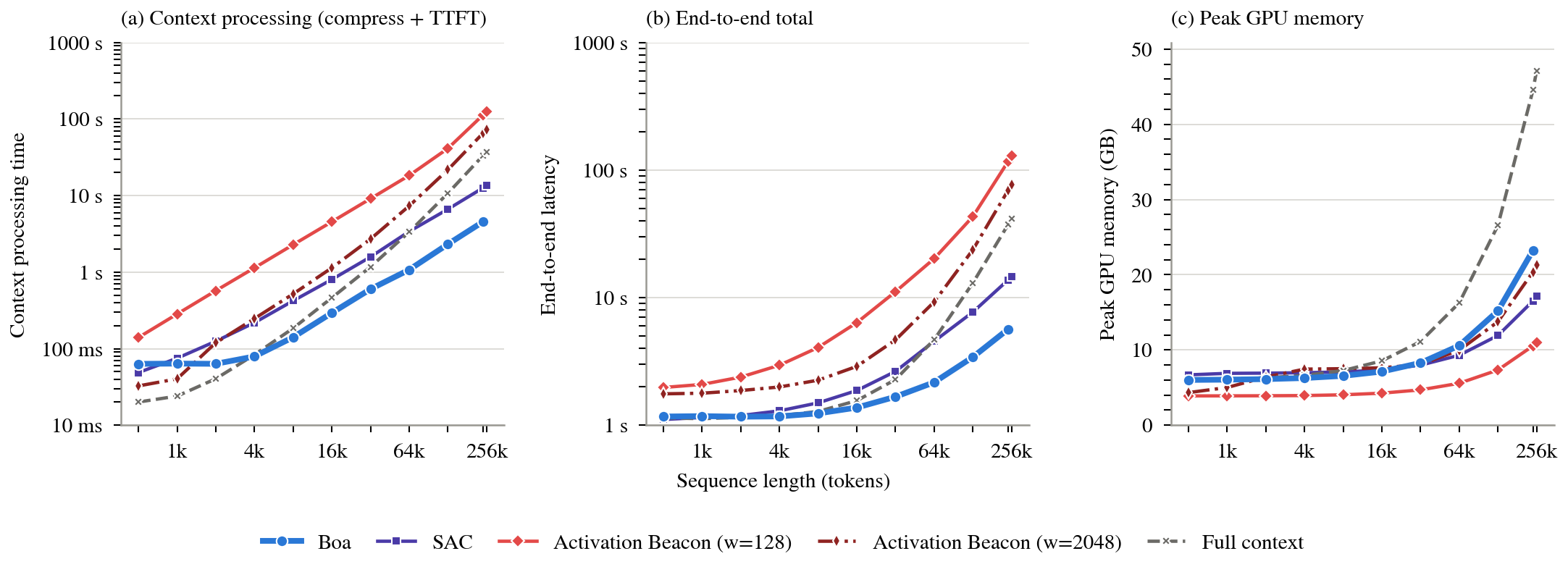}
\caption{Latency and memory comparison at $8\times$ compression with Qwen3-1.7B
(bfloat16, RTX PRO 6000 Blackwell).
\textbf{(a)}~Context processing time ($t_{\text{compress}} + t_{\text{ttft}}$):
elapsed time from raw context to first generated token, log scale.
\textbf{(b)}~End-to-end latency including 64 decoded tokens, log scale.
\textbf{(c)}~Peak GPU memory, linear scale.
Full context (grey dashed) is the uncompressed reference, not a competing
method.}
\label{fig:latency-strip}
\end{figure}

Figure~\ref{fig:latency-strip} profiles wall-clock latency and peak GPU memory
for \methodName{}, SAC, Activation Beacon (window sizes 128 and 2048), and
uncompressed full-context inference, all at $8\times$ compression with a frozen
Qwen3-1.7B decoder in bfloat16 on an NVIDIA RTX PRO 6000 (Blackwell). Each
data point is the median of three timed runs (one discarded warmup), measured
between \texttt{torch.cuda.synchronize()} barriers.

\paragraph{Context processing time (panel~a).}
Different methods split the work between an explicit \texttt{compress()} call
and the decoder's prefill in different ways: SAC runs a separable encoder whose
cost lands entirely in compression time, whereas \methodName{} and Activation
Beacon time their compression inside the decoder forward pass, reporting
zero standalone compression time. To ensure a fair comparison,
panel~(a) plots the sum of compression time and time-to-first-token
($t_{\text{compress}} + t_{\text{ttft}}$), which captures the full elapsed time
from raw context tokens to the first generated token regardless of where each
method draws its internal boundary. \methodName{} remains the
fastest method.

\paragraph{End-to-end latency (panel~b).}
Panel~(b) adds the decode phase (64 generated tokens) to the context processing
cost. Because all methods pay a similar decode overhead, the relative
ordering is preserved.

\paragraph{Peak GPU memory (panel~c).} Panel~(c) reports \texttt{torch.cuda.max\_memory\_allocated} per run. All compression methods remain closely clustered well below the full-context baseline. Activation Beacon with a window of 128 tokens achieves the lowest memory footprint, stemming from its sliding-window design; however, when the window is widened to 2048 for faster inference, its memory consumption rises to the level of \methodName{}. Overall, \methodName{} trades moderately higher
memory for substantially lower latency.

\section{Compression Ratio Scaling: Graceful Degradation}
\label{sec:supp-degradation}

\begin{figure}[!t]
\centering
\includegraphics[width=0.5\linewidth]{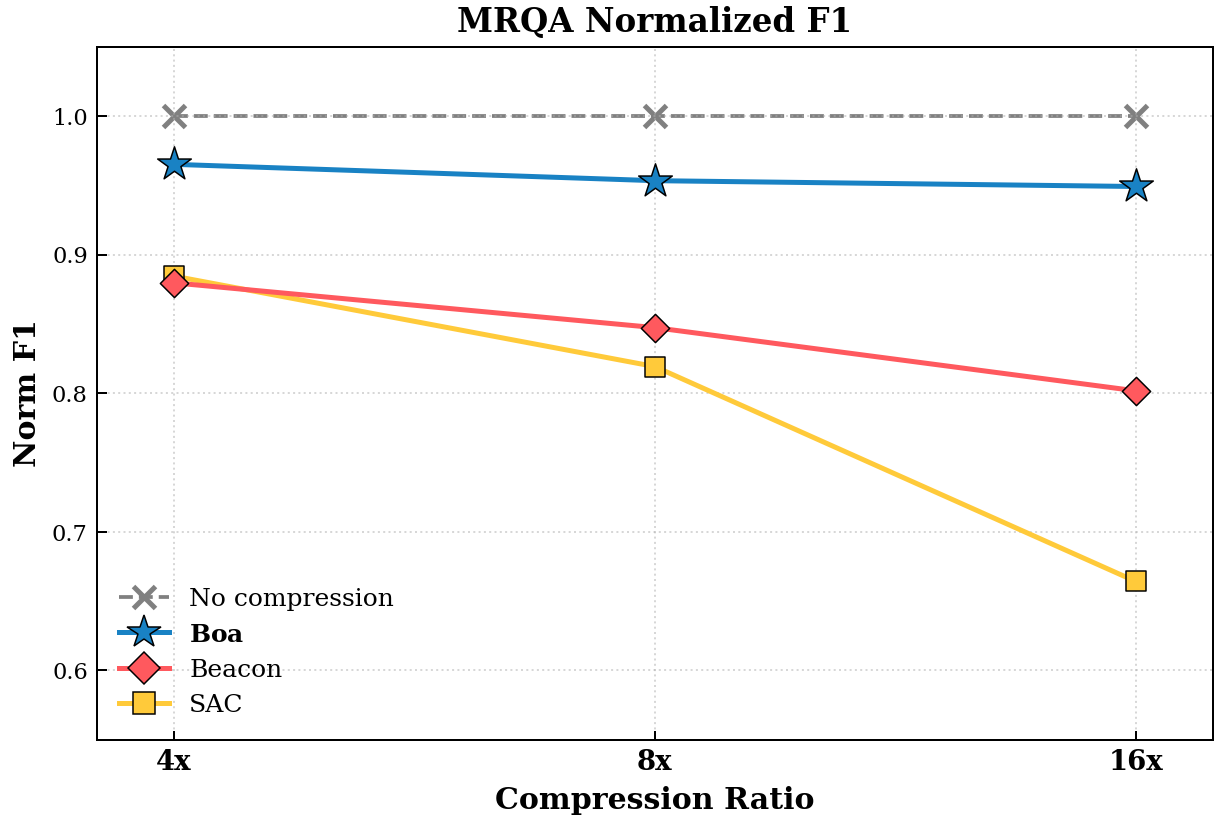}
\caption{In-domain normalised F1 (Norm~$\Delta$) as a function of compression ratio (4$\times$, 8$\times$, 16$\times$) for \methodName{} and Activation Beacon. \methodName{} loses less than 2\% of its normalised F1 across the full range, while Activation Beacon loses roughly 10\%.}
\label{fig:supp-degradation}
\end{figure}

\methodName{} degrades far more
gracefully than the baselines as the compression ratio increases from
4$\times$ to 16$\times$. Figure~\ref{fig:supp-degradation} plots in-domain
normalised F1 (Norm~$\Delta$)
against compression ratio for \methodName{} and Activation Beacon, making the
effect visible directly: \methodName{}'s curve is nearly flat, while
Activation Beacon's drops sharply, and the gap widens as the ratio grows.

\end{document}